\documentclass[runningheads]{llncs}
\usepackage{subfigure}
\usepackage{graphicx}
\usepackage{hyperref}
\usepackage{amssymb}
\usepackage[ruled,linesnumbered]{algorithm2e}
\usepackage{color}
\newcounter{todocnt}

\begin{document}
\title{Long-Term Exploration in Persistent MDPs}
%
%
\author{Leonid Ugadiarov\inst{1} \and Alexey Skrynnik\inst{1,2} \and Aleksandr I. Panov\inst{1,2}\orcidID{0000-0002-9747-3837}}
\authorrunning{L. Ugadiarov et al.}
%
\institute{Moscow Institute of Physics and Technology, Moscow, Russia\\
\and Artificial Intelligence Research Institute FRC CSC RAS, Moscow, Russia\\
\email{skrynnik@isa.ru}}

\maketitle              
%

\begin{abstract}
Exploration is an essential part of reinforcement learning, which restricts the quality of learned policy. Hard-exploration environments are defined by huge state space and sparse rewards. In such conditions, an exhaustive exploration of the environment is often impossible, and the successful training of an agent requires a lot of interaction steps. In this paper, we propose an exploration method called Rollback-Explore (RbExplore), which utilizes the concept of the persistent Markov decision process, in which agents during training can roll back to visited states. We test our algorithm in the hard-exploration Prince of Persia game, without rewards and domain knowledge. At all used levels of the game, our agent outperforms or shows comparable results with state-of-the-art curiosity methods with knowledge-based intrinsic motivation: ICM and RND. An implementation of RbExplore  can be found at  \url{https://github.com/cds-mipt/RbExplore}.

\end{abstract}

\section{Introduction}
Exploration is an essential component of reinforcement learning (RL). During training, agents have to choose between exploiting the current policy and exploring the environment. On the one hand, exploration can make the training process more efficient and improve the current policy. On the other hand, excessive exploration may waste computing resources visiting task-irrelevant regions of the environment~\cite{Burda1}~\cite{Ecoffet}. 

Exploration is essential to solving sparse-reward tasks in environments with high dimensional state space. In this case, an exhaustive exploration of the environment is impossible in practice. A considerable amount of interaction data is required to train an effective policy due to the sparseness of the reward. A common approach is to use knowledge-based or competence-based intrinsic motivation \cite{Oudeyer}. In the first more commonly used approach, it is proposed to augment an extrinsic reward with the additional dense intrinsic reward that encourages exploration~\cite{Bellemare,Burda2,Savinov1}. Another approach is to separate an exploration phase from a learning phase~\cite{Ecoffet}. As noted by the authors of~\cite{Ecoffet}, the disadvantage of the first approach is that an intrinsic reward is a non-renewable resource. After exploring an area and consuming the intrinsic reward, the agent likely will never return to the area to continue exploration due to catastrophic forgetting and inability to rediscover the path because it has already consumed the intrinsic reward that could lead to the area. 

Implementing a mechanism that reliably returns the agent to the neighborhood of known states from which further exploration might be most effective is a challenging task for both approaches. In the case of resettable environments (e.g., Atari games or some robotic simulators), it is possible to save the current state of the simulator and restore it in the future. Many real-world RL applications are inherently reset-free and require a non-episodic learning process. Examples of this class of problems include robotics problems in real-world settings and problems in domains where effective simulators are not available and agents have to learn directly in the real world. Recent work has focused on reset-free setting~\cite{2011.05286,Zhu2020The}. On the other hand, for many domains, simulators are available and widely used at least in the pretraining phase (e.g., robotics simulators~\cite{openai2019learning}).
Specific properties of resettable environments make it possible to reliably return to previously visited states and increase exploration efficiency by reducing the required number of interactions with the environment. Therefore, exploration algorithms should effectively visit all states of an environment. However, factoring in the high dimension of the state space, it is intractable in practice to store all the visited states. Therefore, effective exploration of the environment remains a difficult problem, even for resettable environments.

In this paper, we propose to formalize the interaction with resettable environments as a persistent Markov decision process (pMDP). We introduce the RbExplore algorithm, which combines the properties of pMDP with clustering of the state space based on similarity of states to approach long-term exploration problems. The distance between states in trajectories is used as a feature for clustering. The states located close to each other are considered similar. The states distant from each other are considered dissimilar. Clusters are organized into a directed graph where vertices correspond to clusters, and arcs correspond to possible transitions between states belonging to different clusters. RbExplore uses a novelty detection module as a filter of perspective states. We introduce the Prince of Persia game environment as a hard-exploration benchmark suitable for comparing various exploration methods. The percentage coverage metric of the game's levels is proposed to evaluate exploration. RbExplore outperforms or shows comparable performance with state-of-the-art curiosity methods ICM and RND on different levels of the Prince of Persia environment.
\section{Related Work}
Three types of exploration policies can be indicated. Exploration policies of the first type use an intrinsic reward as an exploration bonus. Exploration strategies of the second type are specific to multi-goal RL settings where exploration is driven by selecting sub-goals. Exploration policy of the third type use clustered representation of the set of visited states.

In recent works~\cite{Burda1,Burda2,pathakICMl17curiosity,Savinov1}, the curiosity-driven exploration of deep RL agents is investigated. The exploration methods proposed by these works can be attributed to the first type. The extrinsic sparse reward is replaced or augmented by a dense intrinsic reward measuring the curiosity or uncertainty of the agent at a given state. In this way, the agent is encouraged to explore unseen scenarios and unexplored regions of the environment. It has been shown that such a curiosity-driven policy can improve learning efficiency, overcome the sparse reward problem to some extent, and successfully learn challenging tasks in no-reward settings.

Another line of recent work focuses on multi-goal RL and can be attributed to the second type. Algorithm HER~\cite{NIPS2017_453fadbd} augments trajectories in the memory buffer by replacing the original goals with the actually achieved goals. It helps to get a positive reward for the initially unsuccessful experience, makes reward signal denser, and learning more efficient especially in sparse-reward environments. A number of RL methods~\cite{NEURIPS2019_57db7d68,NEURIPS2019_83715fd4} focus on developing a better policy for selecting sub-goals for augmentation of failure trajectories in order to improve HER. These policies ensure that the distribution of the selected goals adaptively changes throughout training. The distribution should have greater variance in the early stages of training and direct the agent to the original goal in the latter stages. Other works~\cite{pmlr-v80-florensa18a,Racaniere2020Automated,Skrynnik2019} propose methods to generate goals that are feasible, and their complexity corresponds to the quality of the agent's policy. The distribution of generated goals changes adaptively to support sufficient variance ensuring exploration in goal space.

The Go-Explore~\cite{Ecoffet} algorithm could be attributed to the third type of exploration policy. It builds a clustered lower-dimensional representation of a set of visited states in the form of an archive of cells. Two types of representation are proposed for Montezuma's Revenge environment: with domain knowledge based on discretized agent coordinates, room number, collected items, and without domain knowledge based on compressed grayscale images with discretized pixel intensity into eight levels.

Exploration of the state space is implemented as an iterative process. At each iteration, a cell is sampled from the archive, its state is restored in the environment, and the agent starts exploration with stochastic exploration policy. If the agent visits new cells during the run, they are added to the archive. The statistic of visits is updated for existing cells in each iteration. For both types of representation, the cell stores the highest score that the agent had when it visited the cell. A cell is sampled from the archive by heuristic, preferring more promising cells.

Exploiting domain-specific knowledge makes it difficult to use Go-Explore in a new environment. In our work, we use the idea of clustering of a set of visited states and propose to use a supervised learning model to perform clustering based on the similarity of states. We use a reachability network from the Episodic Curiosity Module~\cite{Savinov1} as a similarity model predicting similarity score for a pair of states. The clusters are organized into a graph using connectivity information between their states in a similar way as the Memory graph~\cite{Savinov2} is built. RND module~\cite{Burda1} is used to detect novel states. Our approach does not exploit domain knowledge, which allows us to apply RbExplore to the Prince of Persia environment without feature handcrafting.

\section{Background}

\subsection{Markov Decision Processes}
A Markov Decision Process (MDP) for a fully observable environment is considered as a model for interaction of an agent with an environment:

\begin{equation} \label{eq:MDP}
    \mathcal{U} = (\mathcal{S}, \mathcal{A}, p, r, \gamma, s_{init}),
\end{equation}
$\mathcal{S}$ --- a state space, $\mathcal{A}$ --- an action space, $p: \mathcal{S} \times \mathcal{A} \times \mathcal{S} \rightarrow \mathbb{R}$ --- a state transition distribution, $r: \mathcal{S} \times \mathcal{A} \times \mathcal{S} \rightarrow \mathbb{R}$ --- a reward function, $\gamma \in \left[ 0; 1 \right]$ --- a discount factor, and $s_{init} \in \mathcal{S}$ --- an initial state of the environment.

An episode starts in the state $s_0$. Each step $t$ the agent samples an action $a_t$ based on the current state $s_t$: $a_t \sim \pi (\cdot \vert s_t)$, where $\pi: \mathcal{S} \times \mathcal{A} \rightarrow \mathbb{R}$ --- a stochastic policy, which defines the conditional distribution over the action space. The environment responds with a reward $r_{t} = r(s_t, a_t, s_{t + 1})$ and moves into a new state $s_{t + 1} \sim p(\cdot \vert s_t, a_t)$. 
The result of the episode is a return $R_{0}$ --- a discounted sum of the rewards obtained by the agent during the episode, where $R_{t} = \sum_{i = t} ^ {T} \gamma ^ {i - t} r_{i} $. Action-value function $Q ^ {\pi}$ is defined as the expected return for using action $a_t$ in a certain state $s_t$: $Q ^ {\pi} (s_t, a_t) = \mathbb{E}_{s_{t + 1} \sim p(\cdot \vert s_t, a_t), a_{t + 1} \sim \pi(\cdot \vert s_{t + 1}) } \left[ R_{t} \vert s_t, a_t \right]$. State-value function $V ^ {\pi}$ can be defined via action-value function $Q ^ {\pi}$: $V ^ {\pi}(s) = \max_a Q ^ {\pi}(s, a)$. The goal of reinforcement learning is to find the optimal policy $\pi ^ *$:

\begin{equation}
    \pi ^ * = \mathrm{argmax}_{\pi} Q ^ {\pi} (s, a) \quad \forall s \in \mathcal{S}, \forall a \in \mathcal{A}
\end{equation}

\subsection{Persistent MDPs}

The persistent data structure allows access to any version of it at any time~\cite{driscoll1989making}. Inspired by that structures, we propose persistent MDPs for RL. We consider an MDP to have a persistence
property if for any state $s_v  \in \mathcal{S}$ exists policy $\pi^p_{s_v}$, which transits agent from the initial state $s_{init}$ to state $s_v$, in a finite number of timesteps $T$. Thus, a persistent MDP is expressed as:

\begin{equation} \label{eq:pMDP}
    \mathcal{U}^{p} = (\mathcal{S}, \mathcal{A}, p, r, \gamma, s_{init}, \pi^p),
\end{equation}
However, the way of returning to visited states can differ. For example, instead of policy $\pi^p_{s_v}$, it could be an environment property, that allows one to save and load states.

\section{Exploration via State Space Clustering}
\begin{figure}[ht]
\centerline{\includegraphics[width=0.7\textwidth]{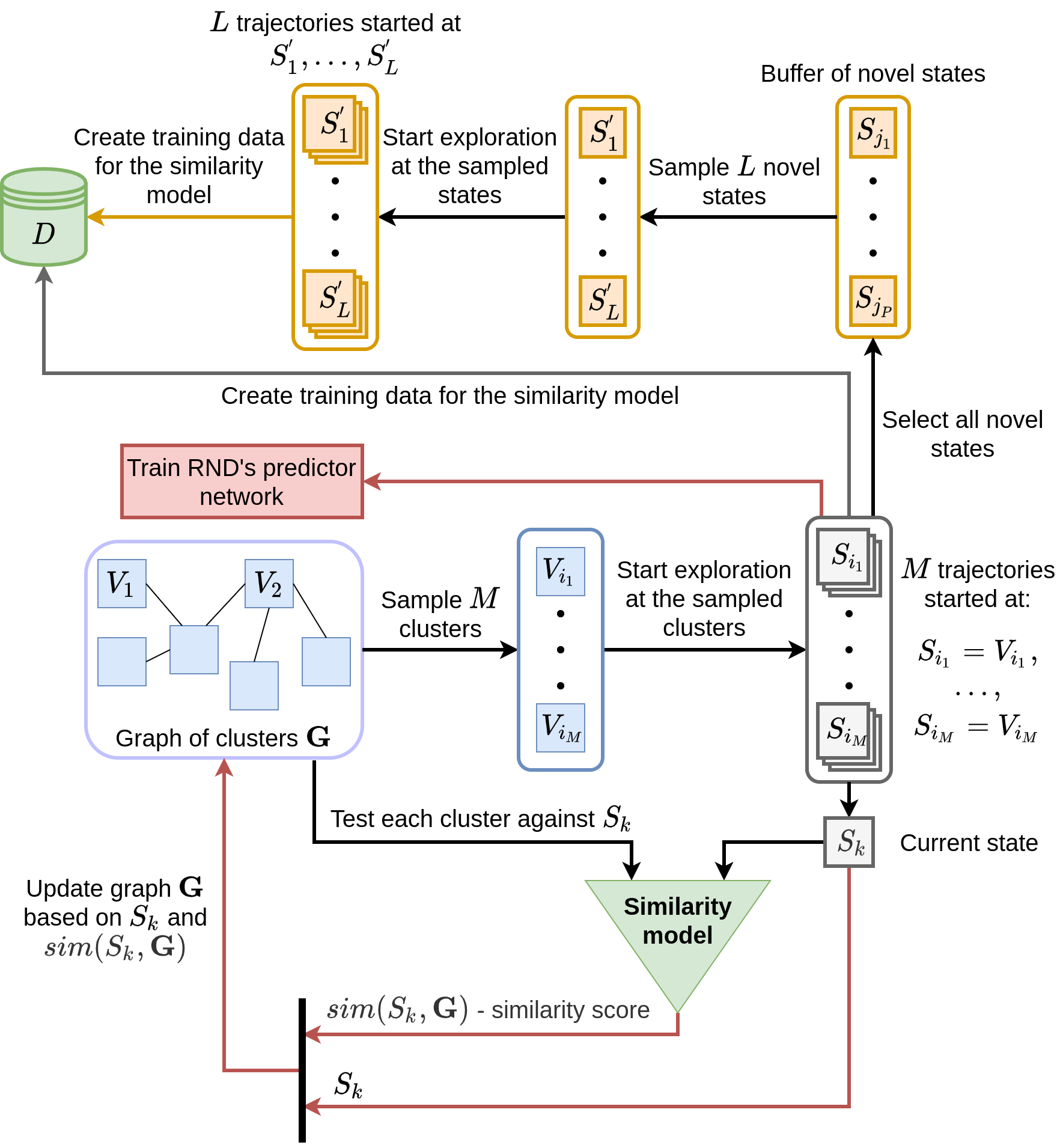}}
\caption{Scheme of the RbExplore algorithm: $M$ exploration trajectories are generated by running exploration from clusters sampled from $\mathbf{G}$. $L$ additional trajectories are generated by running exploration from novel states selected from the exploration trajectories by the novelty detection module. Training data is generated for the similarity model from the exploration trajectories and the additional trajectories. The similarity model is trained on the generated data. $\mathbf{G}$ is updated based on the states from the exploration trajectories and their similarity score with clusters of $\mathbf{G}$. The novelty detection module is trained on the states from the exploration trajectories.}
\label{fig:RbExplore}
\end{figure}

In this paper, we propose the RbExplore algorithm that uses similarity of states to build clustered representation of a set of visited states. There are two essential components of the algorithm: a similarity model, which predicts a similarity measure for a pair of states, and a graph of clusters, which is a clustered representation of a set of visited states organized as a graph. The scheme of the algorithm is shown in Fig.~\ref{fig:RbExplore}.

A high-level overview of one iteration of the RbExplore algorithm:
\begin{enumerate}
    \item Generate exploration trajectories: sample $M$ clusters from the graph of clusters $\mathbf{G}$ based on cluster visits statistics (e.g., preferring the least visited clusters), roll back to corresponding states, and run exploration.
    \item Generate training data for the similarity model $R$ from the exploration trajectories and additional trajectories starting from novel states filtered by the novelty detection module. Full trajectory prefixes are used to generate negative examples.
    \item Train the similarity model $R$.
    \item Update the graph $\mathbf{G}$ with states from the exploration trajectories and merge its clusters. A state is added to the graph $\mathbf{G}$ and forms new clusters if it is dissimilar to states which are already in the graph $\mathbf{G}$. The similarity model is used to select such states.
    \item Train the novelty detection module on the states from the exploration trajectories.
\end{enumerate}
As a result of one iteration, novel states are added to the graph $\mathbf{G}$, the statistics of visits to existing clusters are updated, the similarity model and the novelty detection module are trained on the data collected during the current iteration.

\subsection{Similarity Model}

As a feature for clustering, it is proposed to use the distance between states in trajectories. The states located close to each other are considered similar, the states distant from each other are considered dissimilar. A supervised model is used to estimate the similarity measure between states $R: \mathcal{S} \times \mathcal{S} \rightarrow [0, 1]$. It takes a pair of states as input and outputs a similarity measure between them. The training dataset is produced by labeling pairs of states for the same trajectory $\{\tau ^ k = s_1 ^ k, \ldots, s_T ^ k\}$: triples $(s_i ^ k, s_j ^ k, y_{ij} ^ k)$ are constructed, where $y_{ij} ^ k$ is a class label. States $s_i ^ k, s_j ^ k$ are considered similar ($y_{ij} ^ k = 1$) if the distance between them in the trajectory $\tau ^ k$ is less than $n$ steps: $\vert i - j \vert < n$. Negative examples ($y_{ij} ^ k = 0$) are obtained from pairs of states that are more than $N$ steps apart from each other: $\vert i - j \vert > N$. The model $R$ is trained as a binary classifier predicting whether two states are close in the trajectory (class 1) or not (class 0). 

\begin{figure}[ht]
\centerline{\includegraphics[width=0.9\textwidth]{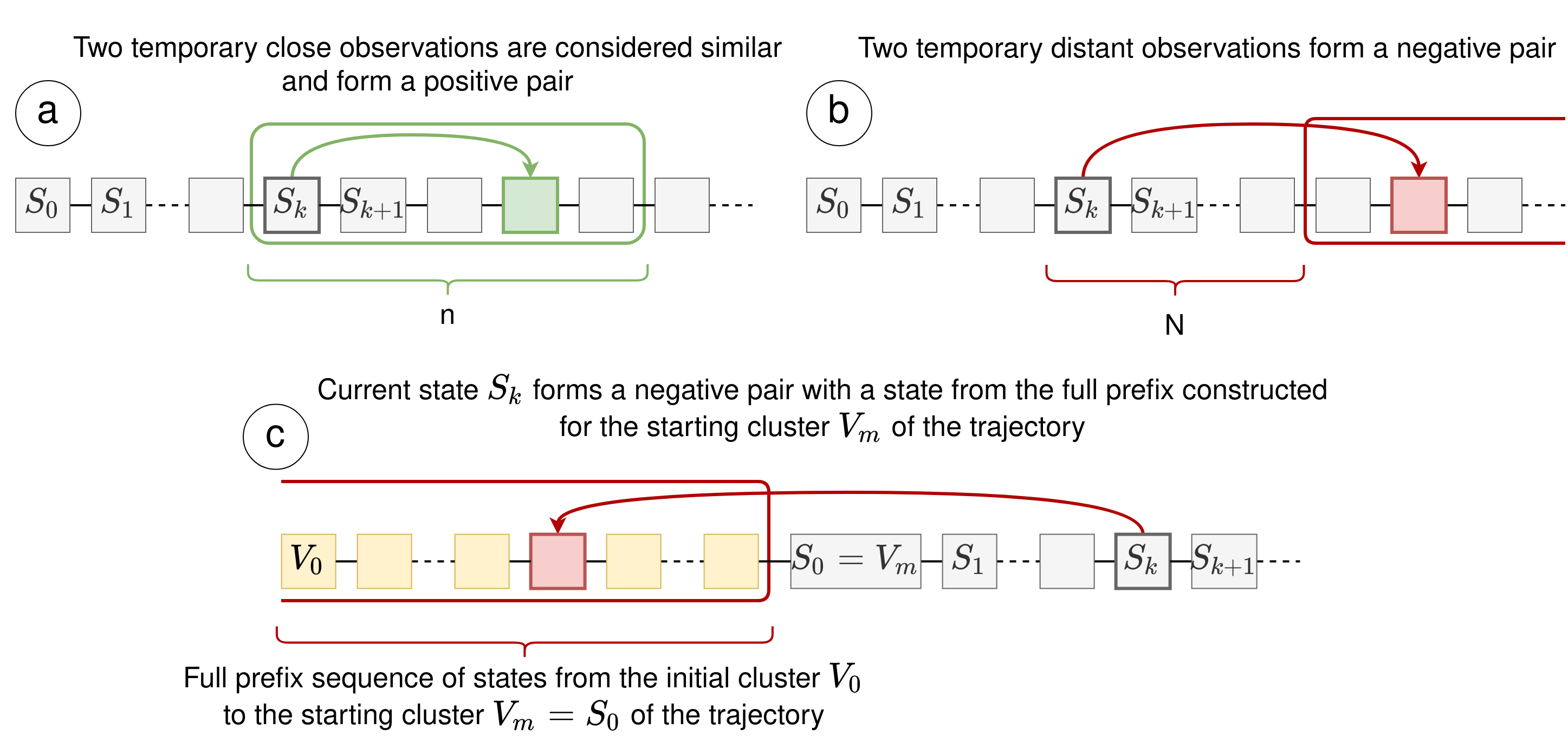}}
\caption{Generation of training data for the similarity model. \textbf{a,b)} Generation of positive and negative examples from states of the same trajectory. \textbf{c)} Generation of negative examples using the full prefix of the starting cluster of the trajectory. }
\label{fig:training-data}
\end{figure}

Fig.~\ref{fig:training-data} illustrates the training data generation procedure. A neural network model is used as a similarity model $R$ as the experiments are performed in environments with high-dimensional state spaces. The network $R$ with parameters $w$ is trained on the training data set $D$ using binary cross-entropy as a loss function:
\[\mathbb{E}_{(s_1, s_2, y) \sim D} \left[-y \log R_w(s_1, s_2) - (1 - y) \log\left(1 - R_w(s_1, s_2)\right)\right] \rightarrow \min_w\]

\subsection{Graph of Clusters}
The clustering of the state space $\mathcal{S}$ is an iterative process using the similarity model $R$ and the chosen stochastic exploration policy $\pi_{explore}(a|s)$ (e.g. uniform distribution over actions). A cluster $v = (s, snap)$ is a pair of state $s$ --- the center of the cluster and the corresponding snapshot of the simulator $snap$. At each iteration, a cluster is selected from which exploration will be continued. The state of the selected cluster is restored in the environment using the corresponding snapshot, and the agent starts exploration with stochastic exploration policy $\pi_{explore}$. For each state $s_i$ of the obtained trajectory $\tau = (s_{1}, snap_{1}), \dots, (s_{T}, snap_{T})$ a measure of similarity with the current set of clusters is calculated. A state $s_i$ is considered as belonging to the cluster $v = (s, snap)$ if the measure of similarity between the state and the cluster's state is greater than the selected threshold $\theta_{sim}$: $R(s, s_i) > \theta_{sim}$. Otherwise, a new cluster is created.

Clusters are organized into a directed graph $(V, E)$. Each vertex of the graph corresponds to a cluster. If two successive states $s_k$ and $s_{k + 1}$ in the same trajectory $\tau = s_1, \dots, s_T$ belong to different clusters $v_i$ and $v_j$, an arc between those clusters $(v_i, v_j)$ is added to the graph. Cluster visit statistics and arc visit statistics are updated each iteration. The graph is initialized with the initial state of the environment $(s_{init}, snap_{init})$. The cluster is selected from the graph for exploration using sampling strategy $\sigma$ that can take into account the structure of the graph and the collected statistics (e.g., the probability of sampling a cluster is inversely proportional to the number of visits). Each iteration of the graph building procedure can be alternated with training the similarity model $R$ on the obtained trajectories. The search for a cluster to which the current state $s_i$ of the trajectory belongs can be accelerated by considering first those vertices which are adjacent to the cluster to which the previous state $s_{i - 1}$ was assigned.

In order to improve the quality of the similarity model $R$ on states from novel regions of the state space $\mathcal{S}$, an RND module is used. For each state, the RND module outputs an intrinsic reward, which is used as a measure of the state's novelty. The state is considered novel if the intrinsic reward is greater than $\beta_{intrinsic}$. At each iteration, all states from the trajectory $\tau$, which are detected by the RND module as a novel, are placed into the buffer of novel states $B$. The buffer $B$ is used to generate additional training data for the similarity model $R$ which includes states from novel regions of the state space $\mathcal{S}$. A set of states $\{(s_i ^ {*}, snap_i ^ {*})\} \sim B$ is randomly sampled from the buffer $B$, and the agent starts an exploration with an exploration strategy by restoring the sampled simulator states. The resulting trajectories are used solely to generate additional training data for the similarity model $R$. 

When an exploration trajectory is processed and a new cluster $v$ is added to the graph $G$ a new arc from $v$ to the parent cluster from which $v$ was created is added to a set of arcs to parent cluster $E_{parent}$. A prefix --- a sequence of states from the parent cluster to $v$ is stored along with the arc. 
Thus, for any cluster in the graph $G$, it is possible to construct a sequence of states that leads to the initial cluster $(s_{init}, snap_{init})$. This property is used to add negative pairs $(s_1, s_2)$ to the training data set of the similarity model $R$ such that $s_1 \in \tau_1$ and $s_2 \in \tau_2$ are distant from each other and $\tau_1 \neq \tau_2$. If the exploration trajectory $\tau$ started from a cluster $(s, snap)$, a full prefix $\tau^{prefix} = s_{init}, \dots, s$ for the trajectory $\tau=(s, snap), \dots$ is constructed to obtain additional negative examples. For a state $s_1 \in \tau$, a sufficiently distant state $s_2 \in \tau ^ {prefix}$ is randomly selected to form a negative example $(s_1, s_2)$. Fig.~\ref{fig:training-data} (c) illustrates this procedure.

Redundant clusters are created at each iteration due to the inaccuracy of the similarity model $R$. A cluster merge procedure is proposed to mitigate the issue. It tests all pairs of clusters in the graph $G$ and merges the pair $v_1 = (s_1, snap_1), v_2 = (s_2, snap_2)$ into a new cluster if the similarity measure of their states is greater than the selected threshold $\theta_{merge}$: $R(s_1, s_2) > \theta_{merge}$. The new cluster is incident to any arc that was incident to the two original clusters. Cluster visits statistics are summarized during merging. As a state and a snapshot of the new cluster, the states and the snapshot of the cluster that was added to the graph $G$ earlier are selected.

\section{The Prince of Persia Domain}

\begin{figure}[ht]
\centerline{\includegraphics[width=0.9\textwidth]{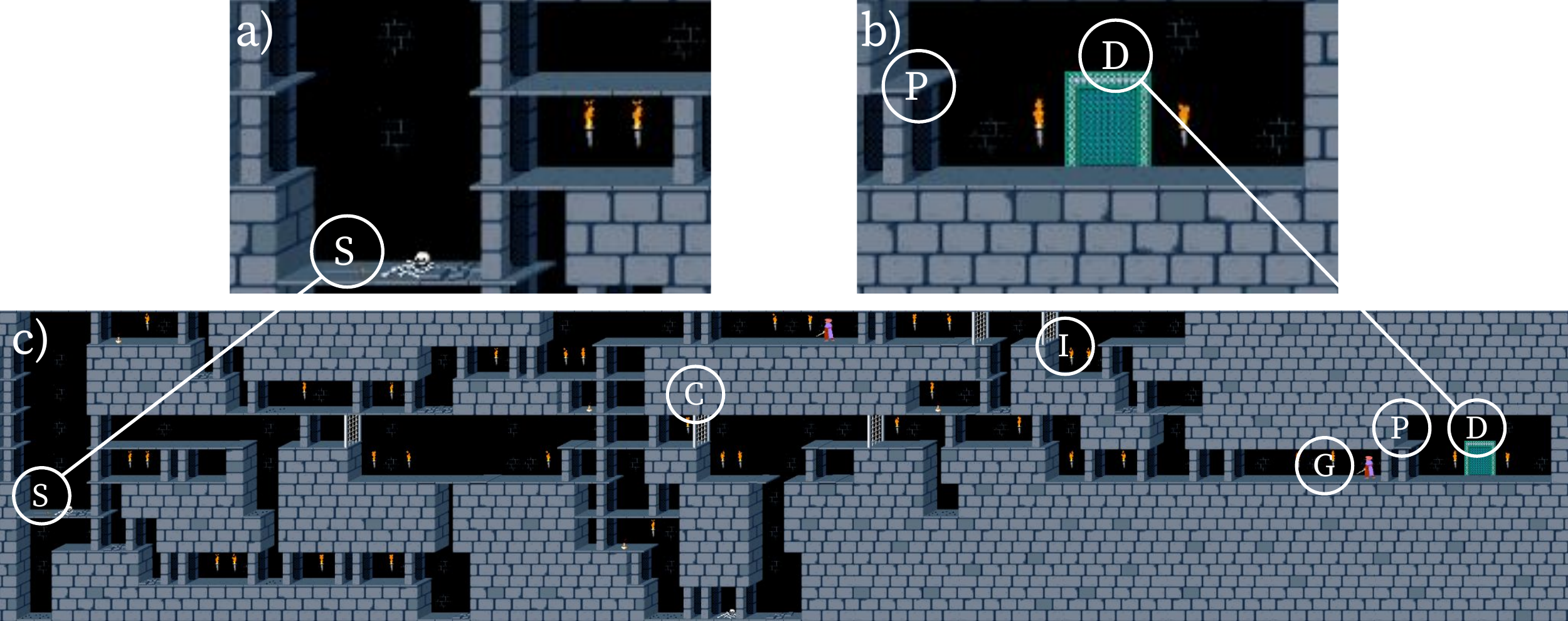}}
\caption{The Prince of Persia environment. \textbf{a,b)} Examples of environment observations from the agent's view. \textbf{c)} The complete map of the first level of the game. The agent's task is to get from the initial location (I) to the final door (D). To solve this problem, the agent needs to: pick up a sword (S), go back and defeat a guard (G), stand on the pressure plate (P) to open the door, proceed to the exit. The environment has many obstacles, such as cell doors (C) and various traps.}
\label{fig:pop-level-one}
\end{figure}

We evaluate our algorithm on the challenging Prince of Persia game environment, which has over ten complex levels. Fig.~\ref{fig:pop-level-one} shows the first level of the game. To pass it, the prince needs to: find a sword, avoiding traps; return to the starting point; defeat the guard, and end the level. In most cases, the agent goes to the next level when he passes the final door, which he also needs to open somehow. 

The input of the agent is a 96x96 grayscale image. The agent chooses from seven possible joystick actions {no-op, left, right, up, down, A, B}. The same action may work differently depending on the situation in the game. For example, the agent can jump forward for a different distance, depending on the take-off run. The agent can jump after using action A or strike with a sword if in combat. Also, the agent can interact with various objects: ledges, pressing plates, jugs, breakable plates. 

The environment is difficult for RL algorithms because it has the action space with changing causal relationships and requires mastery in many game aspects, such as fighting. Also, the first reward is thousands of steps apart from the initial agent position.

We use the percentage coverage metric \% Cov  = $\frac{|U_{visited}|}{|U_{full}|}$, i.e. the ratio of coverage to the max coverage in the corresponding level, where $U_{visited}$ is a set of the visited units and $U_{full}$ full coverage set. We consider the minimum unit to be the area that roughly corresponds to space above the plate. For example, for the first level in the room with the sword, the full area has 36 units, but the agent can visit only 34 of them.

\section{Experiments}

We evaluate the exploration performance of our RbExplore algorithm on the first three levels of the Prince of Persia environment alongside state-of-the-art curiosity methods ICM and RND. 


\subsection{Experimental Setup}
Raw environment frames are preprocessed by applying a frameskip of 4 frames, converting into grayscale, and downsampling to 96x96. Frame pixels are rescaled to the 0-1 range. The neural network $R$ consists of two subnetworks: ResNet-18 network and a four-layer fully connected neural network. ResNet-18 accepts a frame as input and produces its embedding of dimension 512. Embeddings of the pair's frames are concatenated and fed into the fully connected network that performs binary classification of the pair of embeddings.

The same algorithm parameters are used for all levels. The maximum number of frames per exploration trajectory is 1,500. Episodes are terminated on the loss of life.  Training data for the similarity model $R$ is generated with parameters $n=5$ and $N=25$. The similarity threshold $\theta_{similarity} = 0.5$. The similarity threshold for the cluster merging $\theta_{merge} = 0.5$. Merge is run every $15$ iteration. For exploration, clusters are sampled from the graph $G$ with probabilities inversely proportional to the number of visits. Every iteration $M = 30$ clusters are sampled from the graph. The uniform distribution overall actions are used as the exploration policy $\pi_{explore}$. RND module intrinsic reward threshold for detecting novel states $beta_{intrinsic} = 2.5$. $L = 1000$ states are sampled from the buffer of novel states.

To prevent the formation of a large number of clusters at the early stages due to the low quality of the similarity model $R$, the similarity model is pretrained for 500,000 steps with a gradual increase of the value of the similarity threshold from 0 to $\theta_{similarity}$. At the same time, the necessary normalization parameters of the RND module are initialized. After pretraining the graph $G$ is reset to the environment's initial state $(s_{init}, snap_{init})$ and RbExplore is restarted with the fixed similar threshold $\theta_{similairty}$. 


\subsection{Exploring the Prince of Persia Environment}

By design of the Prince of Persia environment, the agent's observation does not always contain information about whether the agent carries a sword. To get around this issue the agent starts the first level with the sword. Also, the agent is placed at the point where the sword is located. This location is far enough from the final door, so reaching the final door is still a challenging task. For the other levels, we did not make any changes to the initial state.

\begin{figure}[ht!]
\centerline{\includegraphics[width=\textwidth]{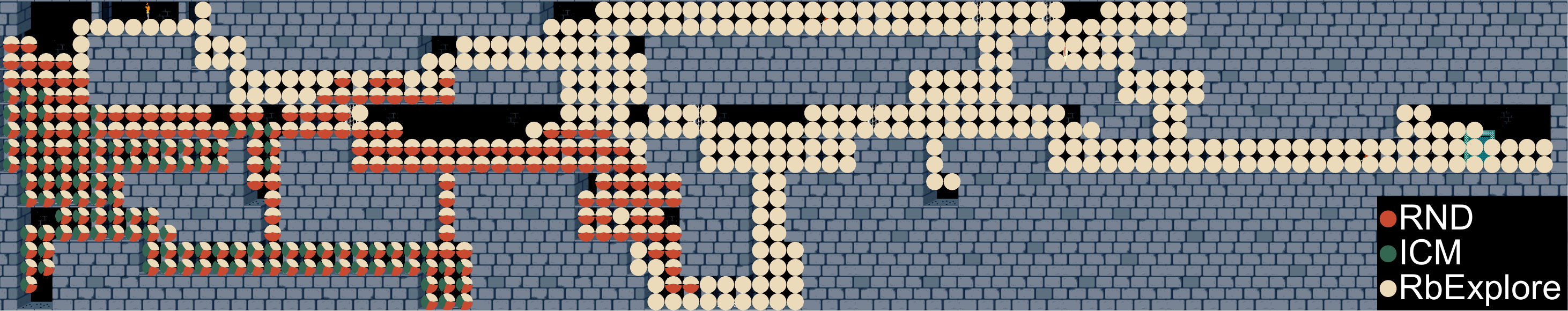}}
\caption{The visualization of the first level coverage for the best RND, ICM, and RbExplore runs. RbExplore significantly outperforms RND and ICM.}
\label{fig:coverage-scheme}
\end{figure}

\begin{figure*}[ht!]
    \centering
    \subfigure{\includegraphics[width=0.32\linewidth]{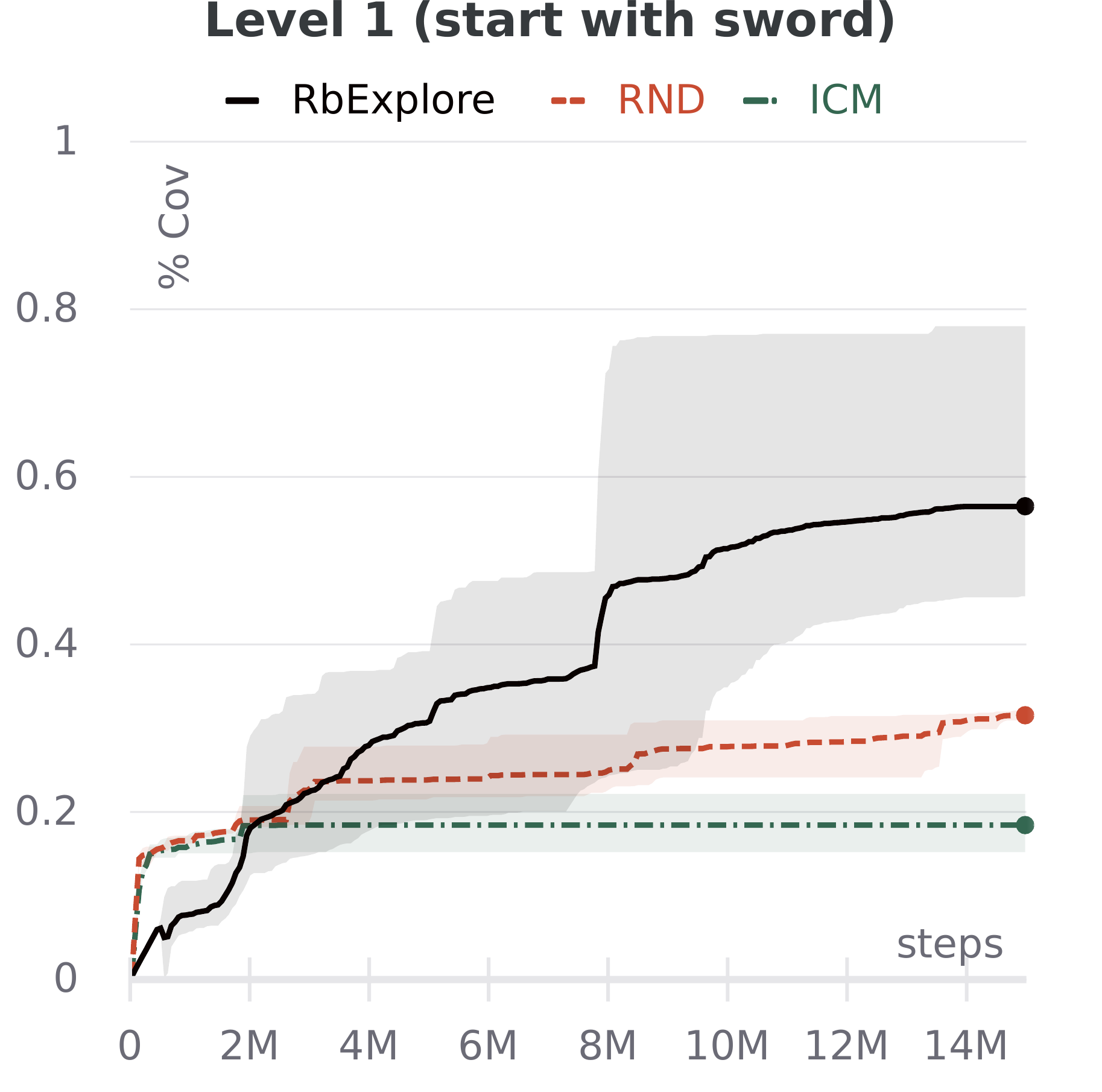}}
    \subfigure{\includegraphics[width=0.32\linewidth]{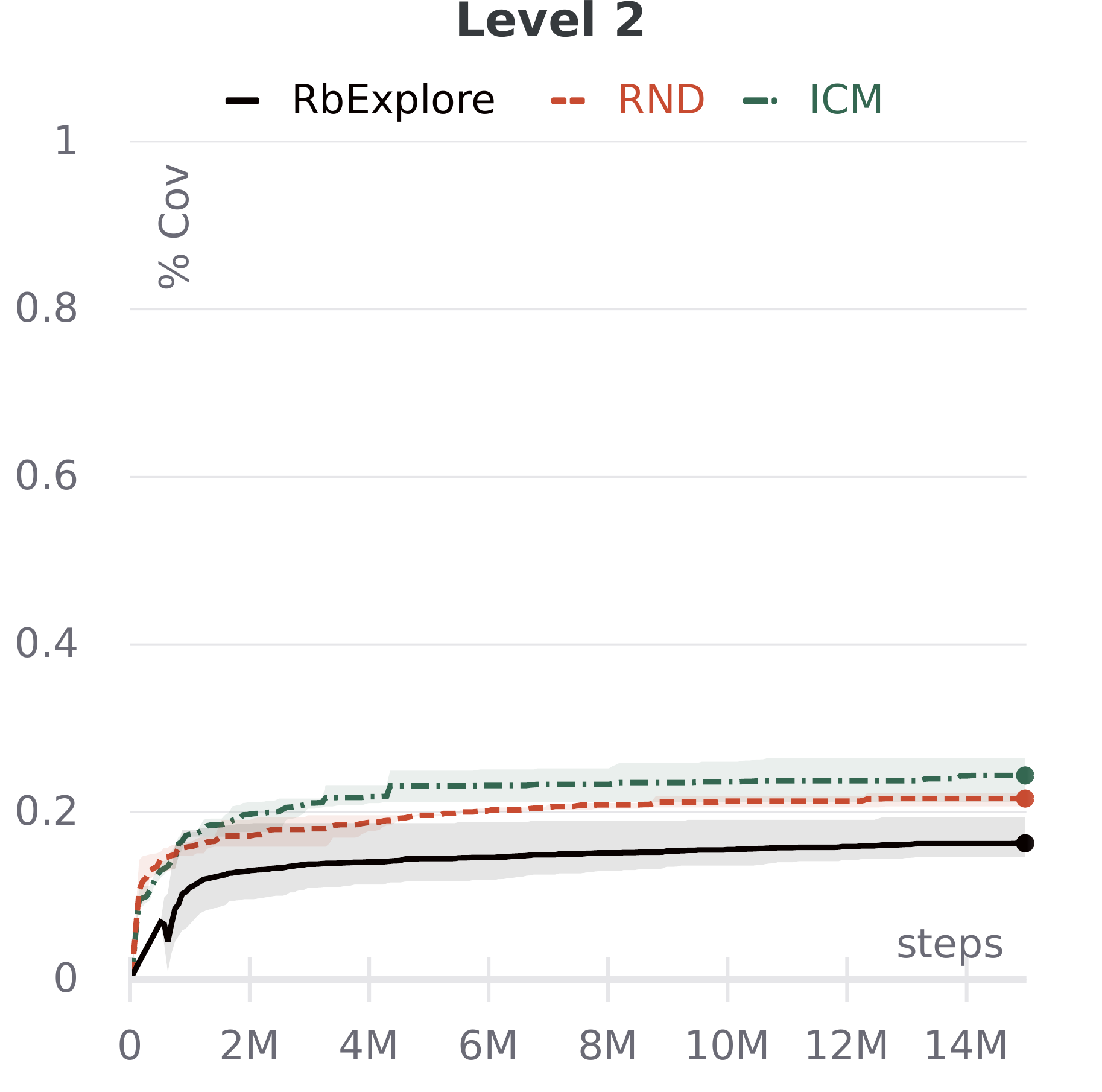}}
    \subfigure{\includegraphics[width=0.32\linewidth]{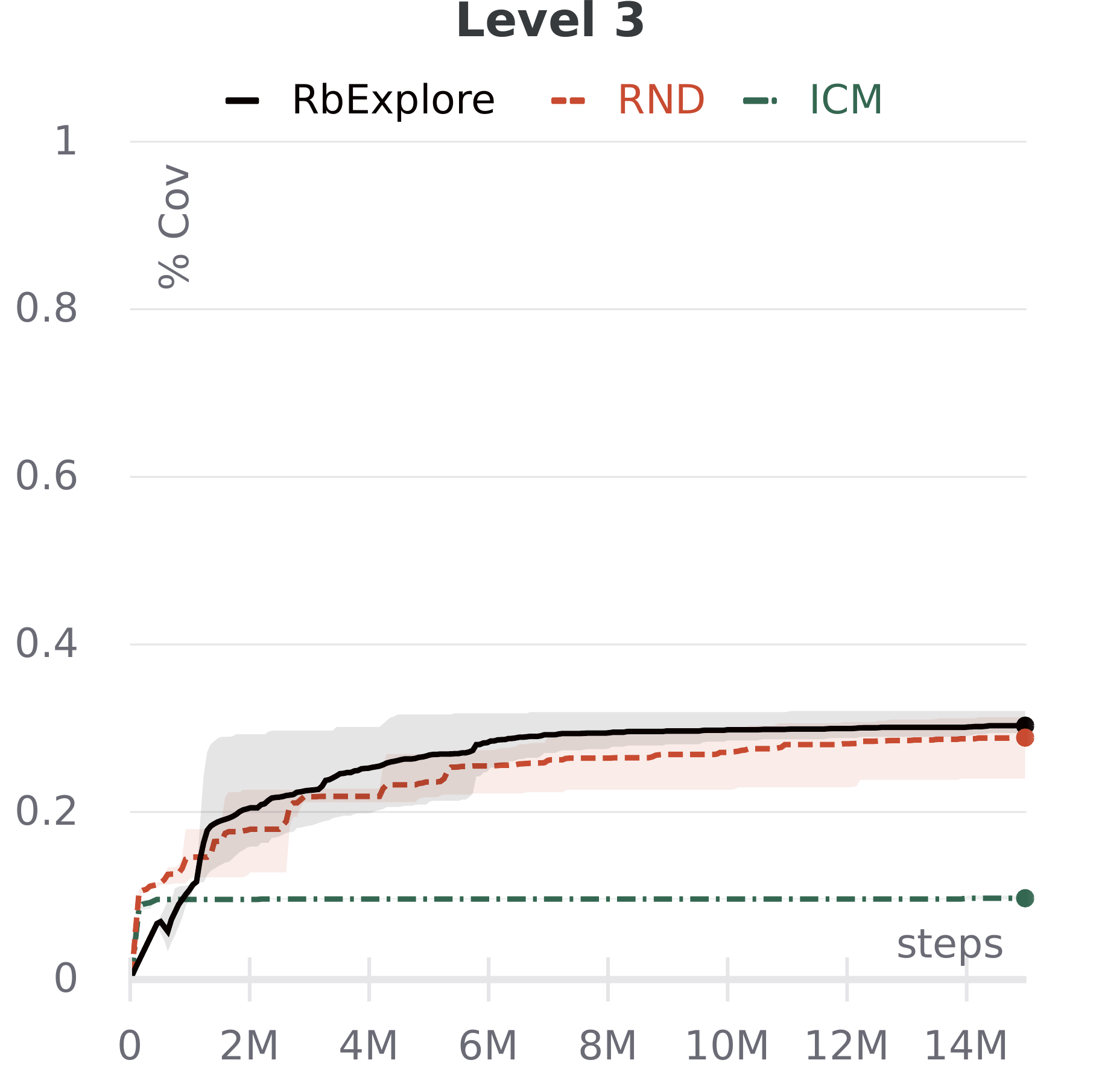}}
    
    \caption{Performance of RbExplore, RND, and ICM for the first three levels of the Prince of Persia environment. \textbf{Left:} Level 1 --- RbExplore significantly outperforms RND and ICM; \textbf{Center:} Level 2 --- All methods resulted in visits to seven rooms. RND and ICM outperform RbExplore as they deal better with local explorations and explore rooms more thoroughly; \textbf{Right:} Level 3 --- RbExplore significantly outperforms ICM and show comparable performance with RND; curves are averaged over three runs. The shading indicates the min-max range.}
    \label{fig:eval-performance}
\end{figure*}

Evaluation of RbExplore, ICM, and RND on the first three levels of the Prince of Persia environment is shown in Fig.~\ref{fig:eval-performance}. On the first level, RbExplore performed significantly outperforms ICM and RND, and also have visited all possible rooms of the level. The visualization of the coverage is shown in Fig.~\ref{fig:coverage-scheme}. 

On levels two and three, none of the algorithms were able to visit all the rooms in 15 million steps. On level two RbExpore shows slightly worse results than RND and ICM. We explain it by the fact that the learning process in RND and ICM is driven by an exploration bonus, which helps them to explore local areas inside rooms more accurately. Each of the algorithms was able to visit only seven rooms at the very beginning of the level. On level three, RbExplore shows slightly better coverage than RND, and both of them outperform ICM.

\subsection{Ablation Study}

\begin{figure*}[ht!]
    \centering
    \subfigure{\includegraphics[width=0.42\linewidth]{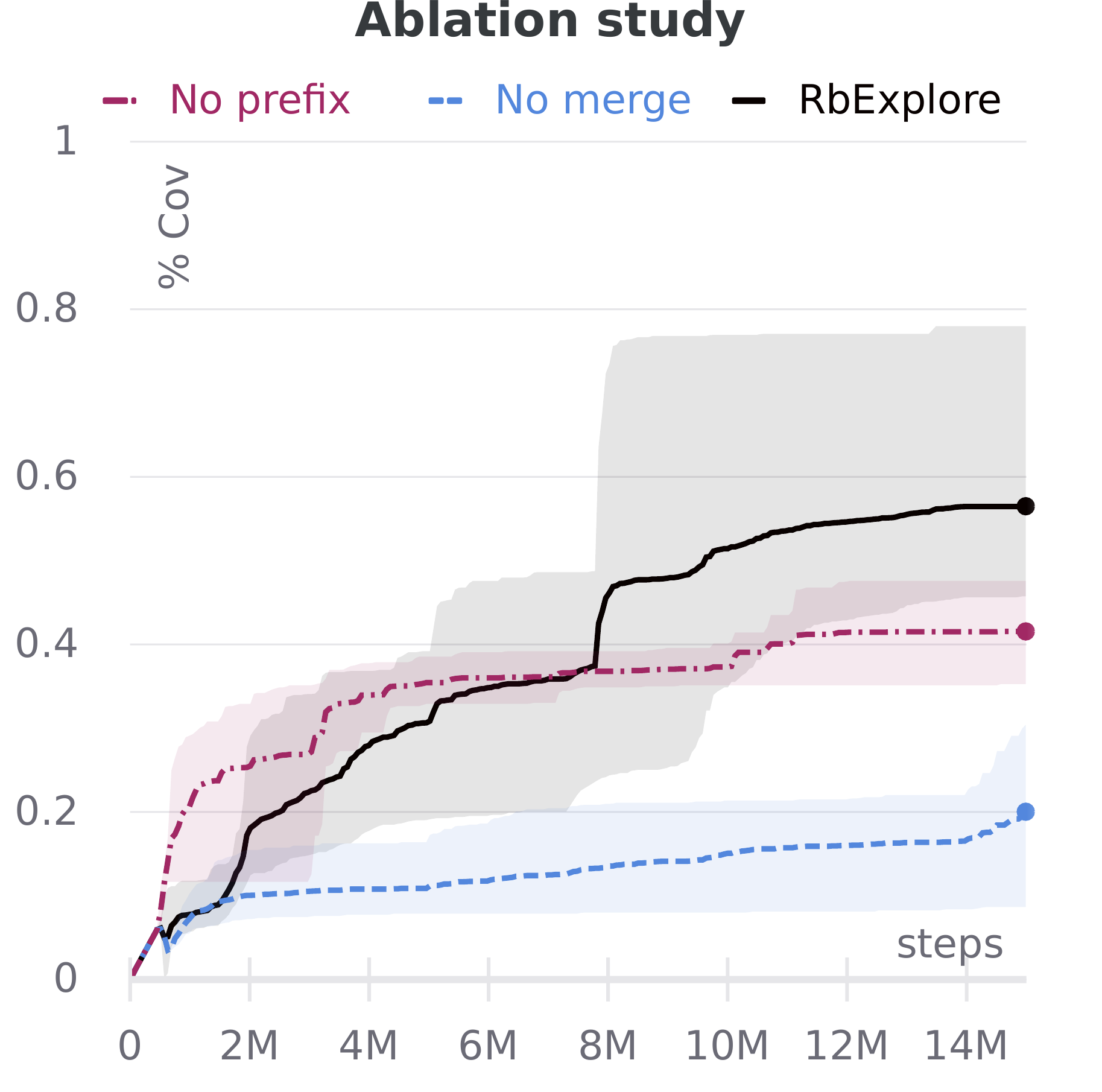}}
    \subfigure{\includegraphics[width=0.42\linewidth]{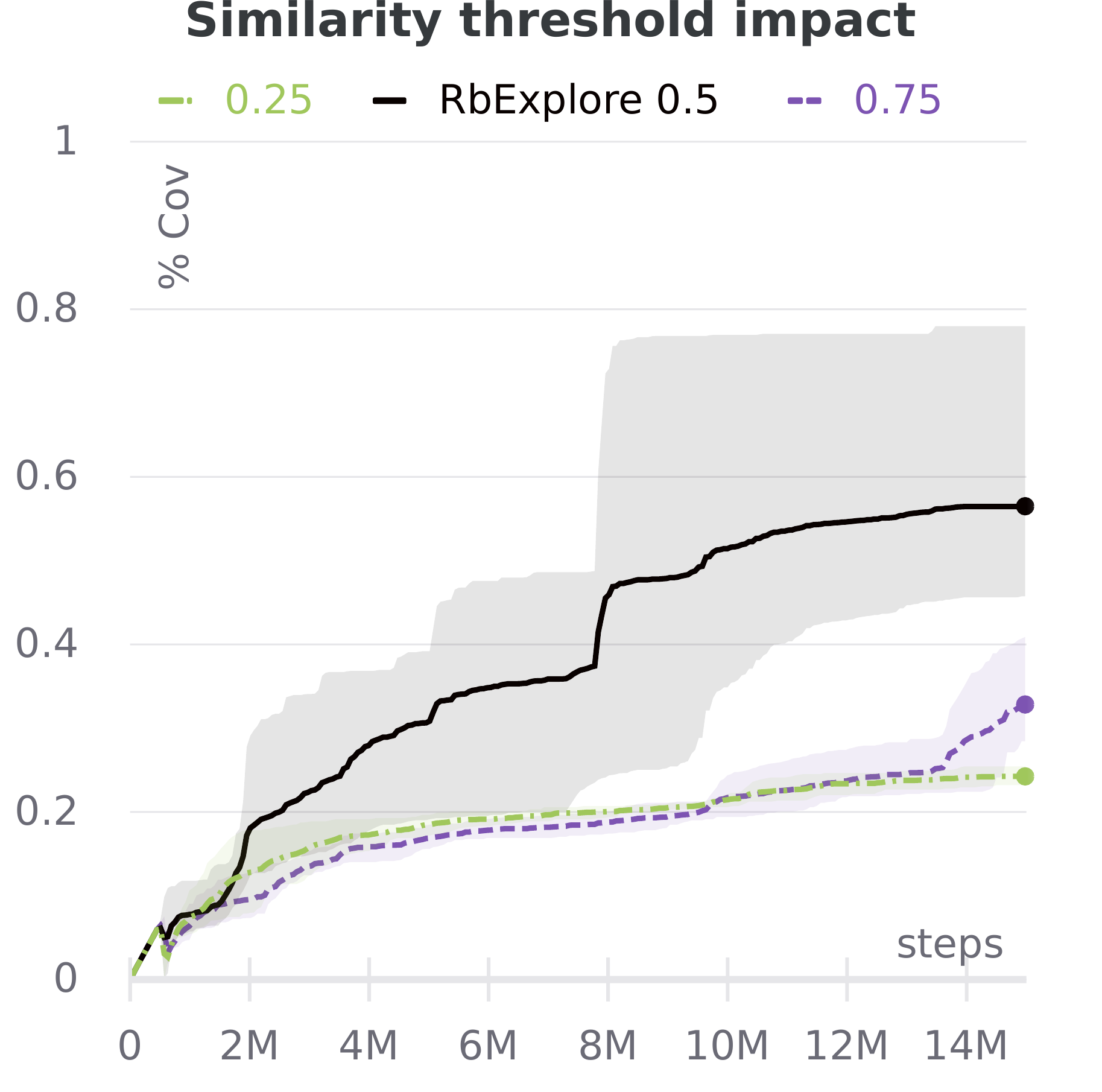}}

    \caption{\textbf{Left:} Level 1 - Comparison of RbExplore with its versions that do not merge clusters (No merge) and do not build the full trajectory prefix to generate a negative example for the training data for the similarity model (No prefix). Disabling merge procedure and generation of negative examples with the use of trajectory prefixes hurts performance of the RbExplore algorithm; \textbf{Right:} Level 1 --- Comparison of RbExplore with similarity thresholds for merging $\theta_{merge} \in \{0.25, 0.5, 0.75\}$. The performance of RbExplore with $\theta_{merge} \in \{0.25, 0.75\}$ is worse than that of RbExplore with $\theta_{merge} = 0.5$; curves are averaged over three runs. The shading indicates the min-max range.}
    \label{fig:ablation}
\end{figure*}

In order to evaluate the contribution of each component of the RbExplore algorithm, we perform an ablation study. Experiments were run on level one with two versions of RbExplore. The first version does not merge clusters in the graph $G$. The second one does not build the full trajectory prefix when generating negative examples for the training data of the similarity model; thus, states of negative pairs are sampled only from the same trajectory. Fig.~\ref{fig:ablation} shows that disabling these components hurts the performance of the algorithm.

Additional experiments were run on level one to study the impact of the value of $\theta_{merge}$ parameter on the performance. Fig.~\ref{fig:ablation} shows that the performance of RbExplore with $\theta_{merge} \in \{0.25, 0.75\}$ is worse than that of RbExplore with $\theta_{merge} = 0.5$.

\section{Conclusion}

In this paper, we introduce a pure exploration algorithm RbExplore that uses the formalized version of a resettable environment named persistent MDP. The experiments showed that RbExplore coupled with a simple exploration policy, which is a uniform distribution over actions, demonstrates performance comparable with RND and ICM methods in the hard-exploration environment of the Prince of Persia game in a no-reward setting. RbExplore, ICM, and RND got stuck on the second and third levels roughly in the same locations where the agent must perform a very specific sequence of actions over a long-time horizon to go further. The combining of RbExplore exploration and exploitation of RL approaches, which also utilize pMDPs, is an important direction for future work to resolve this problem.

\noindent\textbf{Acknowledgements.} 
This work was supported by the Russian Science Foundation (Project No. 20-71-10116).
\bibliographystyle{splncs04}
\bibliography{main}

\begin{thebibliography}{10}
\providecommand{\url}[1]{\texttt{#1}}
\providecommand{\urlprefix}{URL }
\providecommand{\doi}[1]{https://doi.org/#1}

\bibitem{NIPS2017_453fadbd}
Andrychowicz, M., Wolski, F., Ray, A., Schneider, J., Fong, R., Welinder, P.,
  McGrew, B., Tobin, J., Pieter~Abbeel, O., Zaremba, W.: Hindsight experience
  replay. In: Guyon, I., Luxburg, U.V., Bengio, S., Wallach, H., Fergus, R.,
  Vishwanathan, S., Garnett, R. (eds.) Advances in Neural Information
  Processing Systems. vol.~30, pp. 5048--5058. Curran Associates, Inc. (2017),
  \url{https://proceedings.neurips.cc/paper/2017/file/453fadbd8a1a3af50a9df4df899537b5-Paper.pdf}

\bibitem{Bellemare}
Bellemare, M.G., Srinivasan, S., Ostrovski, G., Schaul, T., Saxton, D., Munos,
  R.: Unifying count-based exploration and intrinsic motivation. In: Lee, D.D.,
  Sugiyama, M., von Luxburg, U., Guyon, I., Garnett, R. (eds.) Advances in
  Neural Information Processing Systems 29: Annual Conference on Neural
  Information Processing Systems 2016, December 5-10, 2016, Barcelona, Spain.
  pp. 1471--1479 (2016),
  \url{https://proceedings.neurips.cc/paper/2016/hash/afda332245e2af431fb7b672a68b659d-Abstract.html}

\bibitem{Burda2}
Burda, Y., Edwards, H., Pathak, D., Storkey, A., Darrell, T., Efros, A.A.:
  Large-scale study of curiosity-driven learning. In: ICLR (2019)

\bibitem{Burda1}
Burda, Y., Edwards, H., Storkey, A., Klimov, O.: Exploration by random network
  distillation. In: International Conference on Learning Representations
  (2019), \url{https://openreview.net/forum?id=H1lJJnR5Ym}

\bibitem{driscoll1989making}
Driscoll, J.R., Sarnak, N., Sleator, D.D., Tarjan, R.E.: Making data structures
  persistent. Journal of computer and system sciences  \textbf{38}(1),  86--124
  (1989)

\bibitem{Ecoffet}
Ecoffet, A., Huizinga, J., Lehman, J., Stanley, K.O., Clune, J.: Go-explore: a
  new approach for hard-exploration problems (2021)

\bibitem{NEURIPS2019_83715fd4}
Fang, M., Zhou, T., Du, Y., Han, L., Zhang, Z.: Curriculum-guided hindsight
  experience replay. In: Wallach, H., Larochelle, H., Beygelzimer, A.,
  d\textquotesingle Alch\'{e}-Buc, F., Fox, E., Garnett, R. (eds.) Advances in
  Neural Information Processing Systems. vol.~32, pp. 12623--12634. Curran
  Associates, Inc. (2019),
  \url{https://proceedings.neurips.cc/paper/2019/file/83715fd4755b33f9c3958e1a9ee221e1-Paper.pdf}

\bibitem{pmlr-v80-florensa18a}
Florensa, C., Held, D., Geng, X., Abbeel, P.: Automatic goal generation for
  reinforcement learning agents. In: Dy, J., Krause, A. (eds.) Proceedings of
  the 35th International Conference on Machine Learning. Proceedings of Machine
  Learning Research, vol.~80, pp. 1515--1528. PMLR, Stockholmsmässan,
  Stockholm Sweden (7 2018),
  \url{http://proceedings.mlr.press/v80/florensa18a.html}

\bibitem{openai2019learning}
OpenAI, Andrychowicz, M., Baker, B., Chociej, M., Jozefowicz, R., McGrew, B.,
  Pachocki, J., Petron, A., Plappert, M., Powell, G., Ray, A., Schneider, J.,
  Sidor, S., Tobin, J., Welinder, P., Weng, L., Zaremba, W.: Learning dexterous
  in-hand manipulation (2019)

\bibitem{Oudeyer}
Oudeyer, P.Y., Kaplan, F.: How can we deﬁne intrinsic motivation ? In:
  Proceedings of the 8th International Conference on Epigenetic Robotics:
  Modeling CognitiveDevelopment in Robotic Systems (2008)

\bibitem{pathakICMl17curiosity}
Pathak, D., Agrawal, P., Efros, A.A., Darrell, T.: Curiosity-driven exploration
  by self-supervised prediction. In: ICML (2017)

\bibitem{Racaniere2020Automated}
Racaniere, S., Lampinen, A., Santoro, A., Reichert, D., Firoiu, V., Lillicrap,
  T.: Automated curriculum generation through setter-solver interactions. In:
  International Conference on Learning Representations (2020),
  \url{https://openreview.net/forum?id=H1e0Wp4KvH}

\bibitem{NEURIPS2019_57db7d68}
Ren, Z., Dong, K., Zhou, Y., Liu, Q., Peng, J.: Exploration via hindsight goal
  generation. In: Wallach, H., Larochelle, H., Beygelzimer, A.,
  d\textquotesingle Alch\'{e}-Buc, F., Fox, E., Garnett, R. (eds.) Advances in
  Neural Information Processing Systems. vol.~32, pp. 13485--13496. Curran
  Associates, Inc. (2019),
  \url{https://proceedings.neurips.cc/paper/2019/file/57db7d68d5335b52d5153a4e01adaa6b-Paper.pdf}

\bibitem{Savinov2}
Savinov, N., Dosovitskiy, A., Koltun, V.: Semi-parametric topological memory
  for navigation. In: International Conference on Learning Representations
  (2018), \url{https://openreview.net/forum?id=SygwwGbRW}

\bibitem{Savinov1}
Savinov, N., Raichuk, A., Vincent, D., Marinier, R., Pollefeys, M., Lillicrap,
  T., Gelly, S.: Episodic curiosity through reachability. In: International
  Conference on Learning Representations (2019),
  \url{https://openreview.net/forum?id=SkeK3s0qKQ}

\bibitem{Skrynnik2019}
Skrynnik, A., Panov, A.I.: {Hierarchical Reinforcement Learning with Clustering
  Abstract Machines}. In: Kuznetsov, S.O., Panov, A.I. (eds.) Artificial
  Intelligence. RCAI 2019. Communications in Computer and Information Science.
  vol.~1093, pp. 30--43. Springer (2019). \doi{10.1007/978-3-030-30763-9\_3}

\bibitem{2011.05286}
Xu, K., Verma, S., Finn, C., Levine, S.: Continual learning of control
  primitives: Skill discovery via reset-games (2020)

\bibitem{Zhu2020The}
Zhu, H., Yu, J., Gupta, A., Shah, D., Hartikainen, K., Singh, A., Kumar, V.,
  Levine, S.: The ingredients of real world robotic reinforcement learning. In:
  International Conference on Learning Representations (2020),
  \url{https://openreview.net/forum?id=rJe2syrtvS}

\end{thebibliography}

\end{document}